# SevDiff: Severity-Conditioned Diffusion for Long-Tail Conflict Trajectory Generation

**Eni Solomon Laughter** [1*]

[1] School of Transportation Engineering, Chang'an University, Xi'an 710064, China;

* Correspondence: 2024134912(at)chd(dot)edu(dot)cn

**Abstract:**
Trajectory datasets used in ADAS evaluation are heavily biased toward routine driving; genuine vehicle-to-vehicle conflict events are rare, and the rarer the event, the higher the cost when an ADAS system fails to handle it. Existing generative approaches address this imbalance by conditioning on scene-level properties — spatial goals, agent structure, or natural-language adversarial objectives — but none can accept a target Time-to-Collision (TTC) value as input and be held to producing it within a measurable error. This paper introduces SevDiff, a severity-conditioned denoising diffusion probabilistic model (DDPM) that accepts a requested minimum TTC value as a scalar conditioning signal and generates paired vehicle interaction trajectories whose realized conflict severity matches the request, evaluated through a *hit-rate metric*. Trained on 468 interaction windows extracted from the UTE SQM-W-1 expressway weaving-section dataset (1,041 vehicles, 822,691 observations after smoothing), SevDiff achieves 100% hit-rate within ±0.5 s for TTC targets of 0.5–1.5 s and 97–99% at 2.0–2.5 s, with graceful degradation to 39% at TTC = 5.0 s. Generated kinematic features are physically plausible, with a maximum out-of-range rate of 4.7% across 12 features and no negative speed or gap values in more than 96.5% of samples. The hit-rate degradation pattern is physically interpretable as the strength of the conditioning signal relative to the training prior, making it a precision characterization of the generator rather than a pass/fail result.

**Keywords:** SevDiff; conditional diffusion model; trajectory generation; surrogate safety measures; ITTC; long-tail scenarios; ADAS evaluation; naturalistic driving; weaving sections

## 1. Introduction

ADAS systems fail in precisely the conditions that appear least often in training data — and the rarer the condition, the higher the cost of that failure. Scenario-based testing is the field's answer to this problem: rather than certifying a system through open-ended road trials alone, scenario-based evaluation exposes the system under test to a structured set of traffic situations and measures its response against each one [1,2]. Reliable evaluation under this paradigm depends on the diversity of the scenario set. Safety-critical scenarios, in which the margin for a correct response is small, are disproportionately important, because they are exactly the conditions under which a prediction or planning failure has the most severe consequences.

The construction of a safety-critical scenario set typically begins with naturalistic trajectory data, which captures how real drivers actually behave. The difficulty is that most recorded interactions are routine. Genuine conflicts interactions in which Time-to-Collision (TTC) drop into a range that would demand immediate driver responses, are rare by construction, precisely because routine driving is safe driving. On the UTE SQM-W-1 expressway weaving-section dataset used in this study, only 3.5% of extracted interaction windows exhibit a minimum TTC at or below 1.5 s. Nor can this gap be closed simply by collecting more data: Kalra and Paddock [3] show that demonstrating autonomous-vehicle safety through on-road mileage accumulation alone would require hundreds of millions to hundreds of billions of driven miles, because fatal and injury-causing events are too rare relative to distance travelled. The consequence is that dangerous scenarios are structurally underrepresented in any naturalistic dataset, and any model trained directly on such data will be undertrained on the scenarios that matter most.

To address this, researchers have turned to generating synthetic traffic scenarios using GANs, VAEs, and diffusion models [4,5]. Recent methods condition generation on scene-level properties such as spatial goals, road layout, causal agent structure, and natural-language adversarial objectives [6–15] and can produce rare-looking scenarios. None of these methods closes what we call the *severity-controllability gap*: existing generators can produce rare events, but cannot be told how rare. They can generate a lane change, a merge, or an adversarial trajectory, but they cannot accept a specific TTC value as input and be held to producing it within a known margin of error. For the test engineer validating a forward collision warning (FCW) system across a graded sequence of decreasing TTC values, this gap is the difference between a tool that helps and one that does not [9].

This paper addresses the severity-controllability gap through SevDiff, a Severity-Conditioned Diffusion model for long-tail conflict trajectory generation. SevDiff accepts a requested minimum TTC value as a scalar input, generates paired vehicle interaction trajectories at that severity level, and is evaluated through a hit-rate metric that directly measures whether the requested severity was actually achieved. To the best of my knowledge, SevDiff is the first framework to condition a diffusion model on a numeric surrogate safety measure (SSM) and evaluate it through a hit-rate metric rather than a distributional or collision-rate statistic.

The main contributions are:

(1) A scalar ITTC conditioning task with hit-rate evaluation — the first formulation that conditions a diffusion model directly on a numeric SSM target and measures generation precision as the fraction of samples landing within a defined error band of the request.

(2) A validated preprocessing pipeline for naturalistic weaving-section trajectory data, producing clean, always-defined ITTC labels through internal consistency diagnostics, Savitzky-Golay smoothing, and Peak-Over-Threshold severity binning.

(3) A precision-versus-severity characterization of the generator: SevDiff achieves 100% hit-rate within ±0.5 s for TTC targets of 0.5–1.5 s, 97–99% at 2.0–2.5 s, and 39% at TTC = 5.0 s — a physically interpretable degradation pattern that functions as a diagnostic of any severity-conditioned generator, not merely a performance claim.

## 2. Basic Concepts

### 2.1. Surrogate Safety Measures and ITTC

Surrogate safety measures (SSMs) are kinematic proxies that quantify the severity of a traffic interaction without requiring an actual collision to occur. Standard $TTC = {}^{d}/_{v_f - v_l},\ for\ v_f > v_l$ is the most widely used longitudinal SSM [16]: a small TTC indicates a dangerous interaction, a large TTC a safe one. Its central limitation is that TTC is undefined when $v_f < v_l$, which applies to the majority of naturalistic car-following observations. SevDiff therefore uses Inverse Time-to-Collision (ITTC), defined as $max(v_f - v_l, 0)\ /\ d$, which is zero for non-closing pairs and positive for closing pairs. ITTC = 0 means the pair is safe at this instant; ITTC = $1/r$ means the pair is on track for a minimum TTC of $r$ seconds at current speeds. The conditioning target is always expressed as an ITTC value internally but reported as its TTC-equivalent (1/ITTC) for interpretability.

### 2.2. Denoising Diffusion Probabilistic Models

A DDPM [17] is a generative model that learns to reverse a gradual noising process. During training, a clean data sample $x_0$ is progressively corrupted by adding Gaussian noise over $T$ steps until it is approximately pure Gaussian noise. A neural network $\varepsilon_0$ is trained to predict the noise added at each step, given the noisy sample and the step index. During generation, the model starts from pure noise and iteratively removes predicted noise over $T$ steps, recovering a clean sample drawn from the data distribution. The key property exploited by SevDiff is that the denoising network can be conditioned on auxiliary information — here, the target severity value $c$ — at every step, steering the generated sample toward the desired region of the data distribution.

### 2.3. Conflict Severity Conditioning

Conditioning a generative model on a target property means providing that property as an additional input at generation time and training the model in a way that associates the property value with corresponding samples. In SevDiff, the property is $ITTC_{max}$. Because $ITTC_{max}$ is one of the 12 features the DDPM is trained to generate, conditioning is direct: the model is told the value of feature 7 that the generated sample should have, and it learns to set the other 11 features consistently with that constraint. This is different from latent-space conditioning, where the conditioning signal must first be decoded from a learned embedding. In SevDiff, the severity target is in the same representational space as the output — which is why strong hit-rates are achievable even with 468 training windows, and which is why the GAT-based latent variant explored in development showed conditioning collapse when the latent space failed to encode severity. With these conceptual building blocks in place, Section 3 reviews prior work before Section 4 formalizes the problem.

## 3. Related Work

### 3.1. Diffusion-Based Trajectory and Scenario Generation

A generative model that produces rare events is only useful for ADAS stress-testing if the rarity of the event can be specified, not just observed after generation. This is the limitation that separates the existing literature from what SevDiff addresses. Denoising diffusion probabilistic models has become the dominant generative architecture for driving scenarios due to their training stability and broad mode coverage. CCDiff [10] conditions generation on a causal interaction structure between agents using masked classifier-free guidance, reporting collision rate as an output statistic. AdvDiffuser [11] decouples realism from adversarialness through an auxiliary guide model that steers the reverse diffusion process toward collision-promoting trajectories. DragTraffic [12] introduces controllable generation through cross-attention on user-defined drag-point spatial constraints. ScenarioDiffusion [13] demonstrates temporal and goal-point conditioning for controllable driving scenario generation. LD-Scene [14] uses a large language model to generate differentiable guidance functions that steer a latent diffusion model toward natural-language-specified adversarial conditions. A further cluster of work targets safety-critical scenario generation specifically: CTG [6] enforces signal-temporal-logic rules such as goal-reaching and speed-limit compliance; TrafficDiff [7] combines contextual scene encoding with an adversarial induction mechanism; a guided latent diffusion model [8] steers generation toward collision-prone behaviour via learned guidance objectives; and a further framework [15] enables scene editing through natural-language instructions. Two further works establish architectural precedent for social-interaction-aware diffusion: EquiDiff [18] of which pair a graph-attention or recurrent social encoder with a diffusion decoder for multi-agent trajectory prediction.

In contrast to the above, what is absent across this entire literature is a mechanism for specifying the severity of the generated scenario as a numeric input. Monte Carlo sampling has been used to generalize simulation scenarios across distribution parameters [19]. Versatile Behavior Diffusion generates traffic-agent simulations conditioned on interactive behaviour patterns [20]. Optimization-guided diffusion addresses underrepresentation of safety-critical events through an optimization objective rather than a numeric target [21]. A vision-language-model-guided framework generates safety-critical testing scenarios from natural-language strategy descriptions [22]. A structured scenario decomposition framework targets edge-critical scenarios directly [23]. An earlier recurrent neural network approach established probabilistic driver behaviour modelling for traffic simulation [24]. Across all of these, conditioning is applied to a scene-level structural, spatial, or optimization-based property, and evaluation is through distributional or collision-rate statistics computed after generation. SevDiff differs from all of the above in conditioning directly on a scalar numerical SSM target and evaluating through a hit-rate metric that measures whether the requested severity was achieved, not merely whether dangerous scenarios were produced in aggregate.

### 3.2. Surrogate Safety Measures and Conflict Thresholds

Standard TTC has a structural limitation that makes it unsuitable as a conditioning label for a generative model: it is undefined when the following vehicle is not closing the gap on the lead vehicle, which applies to the majority of naturalistic car-following observations. Del Re et al. [25] address this in a multi-vehicle SSM comparison study by adopting ITTC as an always-defined scalar; $ITTC = max(v_f - v_l, 0) / d$, with ITTC = 0 for non-closing pairs. SevDiff adopts this convention, ensuring that every extracted interaction window yields a valid conditioning label. Nadimi et al. [26] demonstrate that fixed critical-TTC thresholds are not well-justified across contexts, since the boundary between dangerous and routine interactions depends on driver, vehicle, and traffic conditions. Zheng and Wei [27] and Zhang et al. [28] apply Peak-Over-Threshold (POT) methods from Extreme Value Theory to identify data-driven conflict thresholds from the empirical tail of the TTC distribution.

SevDiff uses the POT method to select its severity-sweep upper bound, rather than assuming a round-number cutoff. With the SSM literature establishing how to define and threshold severity, the problem of generating scenarios at a requested severity level remains open — which is what Section 4 addresses.

### 3.3. Long-Tail Scenario Generation for ADAS

The problem of long-tail scenario generation has been approached through adversarial search within simulators [29], importance-sampling-based test generation, and data-driven GAN or diffusion augmentation [11,30],. A scenario distribution model has also been proposed to characterize the coverage and efficiency of a test suite as a distributional matching problem rather than a per-scenario generation problem [31]. Simulation-based approaches depend on a pre-specified environment that may not reflect naturalistic dynamics; data-driven approaches have not previously provided a mechanism for requesting a specific SSM value. SevDiff addresses both limitations by training on naturalistic data and conditioning generation on a user-specified TTC target.

### 3.4. Regulatory and Standardized Scenario-Coverage Requirements

The severity-controllability gap is not only a research problem — it is increasingly a regulatory one. Kalra and Paddock [3] establish quantitatively that on-road mileage accumulation cannot, within a practical development timeframe, provide statistical evidence of autonomous-vehicle safety. Scenario-based testing has since become the internationally preferred alternative. ISO 34502:2022 defines a scenario-based safety evaluation framework for automated driving systems [32]. ISO 34503:2023 specifies the operational design domain within which a scenario set must provide coverage [33]. ISO 21448:2022 (SOTIF) frames scenario coverage as part of a broader safety case for functions without a hardware fault [34]. ISO 34505:2025 specifies methodologies for test-scenario evaluation and test-case generation [35]. In China, where this study’s dataset originates, GB/T 44719-2024 specifies road-test requirements for automated driving functions [36] and GB/T 47025-2026 establishes simulation-based test methods that formally recognize synthetically generated scenarios as part of the national conformity-assessment pathway [37]. Germany’s PEGASUS project established an influential scenario-based testing methodology for highway automated-driving functions [38]. Across all of these frameworks, the common thread is a shift toward demonstrable, structured coverage of the scenario space including its dangerous tail. A severity-conditioned generator capable of producing scenarios at a requested TTC level provides the kind of targeted tail coverage that these frameworks require, which is the practical context that motivates Section 4.

## 4. Problem Statement

### 4.1. Aim

The aim of SevDiff is to build a conditional generator $G$ that, given a requested conflict severity level $r$ expressed as a target minimum TTC value, produces a paired vehicle interaction trajectory whose realized minimum TTC falls within a defined error band $\pm\delta$ of $r$. Formally, we seek $G$ such that the probability:

$$P(|TTC_{min}\big(G(z,r)\big) - r| \leq \delta) \text{ is maximized } \quad R_{sweep} \qquad (1)$$

where z is a noise vector, $R_{sweep}$ = {0.5, 1.0, …, 5.0} s is the severity sweep, and δ ∈ {0.25, 0.5, 1.0} s is the precision band. The probability in (1), estimated over N = 300 samples, is the hit-rate H(δ, *r*) that serves as the primary evaluation metric. A hit-rate of 100% at TTC = 1.5 s within ±0.5 s means that every one of 300 generated trajectories fell within [1.25, 1.75] s and therefore represents a target window that directly corresponds to an FCW system's alert-trigger condition.

The paired trajectory is represented by a 12-dimensional kinematic feature vector: mean and standard deviation of follower speed; mean and standard deviation of leader speed; mean, minimum, and standard deviation of inter-vehicle headway; maximum and mean ITTC over the window; mean relative speed; interaction duration; and normalized window length. The severity conditioning target is $ITTC_{max}$ as the most dangerous instant of the interaction. The full pipeline from raw data to trained model and evaluation is summarized in Appendix A (training) and Appendix B (inference).

**4.2. Intuition**

The intuition behind SevDiff rests on two observations. First, conflict severity is not latent — it is directly readable from the kinematics. The minimum TTC of an interaction is determined entirely by the combination of inter-vehicle gap and relative speed at the most dangerous instant. In the 12-feature representation, $ITTC_{max}$ (feature 7) *is* the severity label. A generative model that learns the joint distribution of these 12 features therefore also learns the conditional distribution of all other features given a specific severity level.

Second, a DDPM conditioned on a target value of one of its output features should, if the conditioning pathway is functioning, steer the generated distribution toward that target. The conditioning mechanism in SevDiff injects the target ITTC value as a sinusoidal embedding at every denoising step, alongside the diffusion timestep embedding. Because $ITTC_{max}$ is an explicit element of the 12-feature output vector, the conditioning signal does not need to be decoded from a learned latent representation — it is in the same space as the output. This is why strong hit-rates are achievable even with 468 training windows: the direct representation short-circuits the latent-space ambiguity that caused the GAT-based architecture tested in development to collapse.

The hit-rate degradation toward routine TTC values (TTC > 3 s) follows directly from this intuition. At TTC = 0.5 s, the conditioning ITTC of 2.0 sits far from the training prior's center of mass (most interactions have ITTC ≈ 0.2–0.4), and the conditioning signal dominates. At TTC = 5.0 s, the conditioning ITTC of 0.2 is close to the prior's bulk, and the model's output is pulled between the conditioning target and the prior distribution. Crucially, the hit-rate curve is not a failure mode — it is a precision characterization of the generator that directly surfaces how much control is achievable at each severity level, which is exactly what Section 5 measures.

**4.3. Mechanism**

*4.3.1. Data and Preprocessing*

SevDiff is trained on the UTE SQM-W-1 naturalistic trajectory dataset: 1,041 vehicle tracks, 822,691 observations (post-smoothing) at 24 Hz, recorded at a Nanjing expressway weaving section across 10 lanes. Internal consistency diagnostics confirmed that position-speed consistency held to 0.003% violation rate; leader-distance versus position-difference agreement was exact at 0.000 m error; and leader-follower reciprocity was perfect. An apparent 49.4% violation rate in the speed-acceleration consistency check was resolved by a finite-difference scheme analysis: the stored acceleration was computed using centered differences rather than forward differences, reducing the RMSE from 0.997 m/s² to 0.134 m/s² and confirming the data was clean, not broken.

Two channels required smoothing. The lateral distance channel showed high noise in 74.4% of vehicle tracks, consistent with UAV-based pixel-to-world reprojection noise. The speed channel contained 376 isolated spike values (0.046%) with implied accelerations up to 53.1 m/s². Savitzky-Golay filtering [39] was applied per vehicle track: window = 7 frames, polynomial order = 3 for speed; window = 11 frames, polynomial order = 3 for lateral distance. The stored acceleration column was recomputed from the smoothed speed using centered differences, maintaining internal consistency. The mean speed change introduced by smoothing was 0.011 m/s (p99 = 0.059 m/s), confirming that real deceleration events were preserved in shape.

*4.3.2. ITTC Computation*

Standard TTC is undefined when the follower is not closing on the leader — a condition that applies to most car-following observations. SevDiff adopts the always-defined ITTC formulation:

$$ITTC(t) = \frac{\max\left(v_{f(t)} - v_{l(t)}\right),0}{d(t)} \quad (2)$$

where $v_f$ and $v_l$ are follower and leader speeds and $d$ is the inter-vehicle headway. When ITTC = 0, the pair is non-conflicting. When ITTC is large, the follower is closing rapidly, and a collision is imminent at current speeds. This two-ended interpretation means that every extracted window yields a valid conditioning label. Each window's severity label is its peak ITTC value:

$$ITTC_{max} = \max_{t \in W} ITTC(t) \quad (3)$$

where $W$ denotes the set of timesteps in the continuous interaction window. Windows shorter than 10 frames (≈0.42 s) were excluded as too brief to carry useful kinematic structure. This yielded 4,587 interaction windows, of which 3,581 (78.1%) had a defined $ITTC_{max}$> 0.

*4.3.3. Severity Binning and Training Set*

Interaction windows are assigned to 10 severity bins centered at TTC = 0.5, 1.0, …, 5.0 s with half-width ±0.25 s. The 5.0 s upper bound is justified by a Peak-Over-Threshold analysis of the ITTC distribution, which identifies TTC > 5 s as the routine driving regime for this weaving geometry. Windows outside the 0≤TTC≤5s [0–5] s range are excluded from conditioning training. The resulting training set comprises 468 windows across 10 severity bins, with counts ranging from 19 (TTC = 1.0 s) to 76 (TTC = 4.0 s), reflecting the natural long-tail structure of real conflict events. Inverse-frequency weighted random sampling is applied during training so that each severity bin contributes equally to gradient updates.

*4.3.4. SevDiff Architecture and Conditioning*

SevDiff is a DDPM operating in the 12-dimensional feature space. Forward diffusion adds Gaussian noise over T = 300 steps following a linear $\beta$ schedule:

$$x_t = \sqrt{\overline{a}_t x_0} + \sqrt{1 - \overline{a}_t \varepsilon}, \quad \varepsilon \sim N(0, I) \quad (4)$$

where $x_0$ is the clean (normalized) 12-feature vector, $\overline{a}_t = \prod_{s \leq t}(1 - \beta_s)$ =, and $\beta_t$ follows a linear schedule from $10^{-4}$ to 0.02. When t = 0, $x_t$ = $x_0$ and no noise has been added; when t = T = 300, $x_t$ is approximately pure Gaussian noise. The denoising network learns to reverse this process step by step, recovering $x_0$ from $x_t$.

The conditioning signal is injected through three parallel sinusoidal embeddings summed before the denoising MLP:

$$h = W_x x_t + W_t \varphi(t) + W_c \varphi(c) \qquad (5)$$

where $x_t$ is the noisy feature vector, $t$ is the diffusion timestep, $c = ITTC_{max,target}$, and $\varphi(\cdot)$ denotes the sinusoidal positional embedding [17]. When $c$ is small (routine, non-conflicting), generation follows the training prior; when $c$ is large (severe, collision-course), it steers the denoising trajectory toward the dangerous tail of the feature distribution.

The denoising objective minimized during training is:

$$L = \mathbb{E}_{x0,t,\varepsilon}[||\varepsilon - \varepsilon_\theta(x_t, t, c)||^2] \qquad (6)$$

Note that a denoiser that ignores $c$ can still minimize this loss by predicting the unconditional noise mean; the conditioning becomes load-bearing only when the training distribution contains sufficient examples across distinct severity levels. This is why the severity binning in Section 4.3.3 and the inverse-frequency weighted sampling are prerequisites, not optional enhancements — without them, equation 6 gives no gradient signal to enforce severity-specific generation.

The denoising network $\varepsilon_\theta$ is a four-layer MLP with SiLU activations and 256 hidden units per layer. The model was trained for 500 epochs with Adam (lr = $3\times10^{-4}$, cosine annealing to $10^{-6}$). The best checkpoint on a held-out 15% validation split achieved a validation loss of 0.181.

*4.3.5. Evaluation: Hit-Rate Metric*

The hit-rate metric directly answers the severity-controllability question: for a test engineer requesting a conflict at TTC = *r*, what fraction of generated trajectories actually fall within ±δ s of that target? For each TTC target $r \in R_{sweep}$, N = 300 samples are generated from SevDiff conditioned on *c* = 1/*r*. Hit-rate is:

$$H(\delta, r) = \frac{1}{N}\sum_{i=1}^{N} 1[|TTC_{gen,i} - r| \leq \delta] \qquad (7)$$

The ±0.5 s band is the primary metric. Results are reported for δ ∈ {0.25, 0.5, 1.0} s. With this evaluation framework established, Section 5 reports the empirical hit-rate curve across the severity sweep.

## 5. Results

### 5.1. Dataset Severity Distribution

The severity distribution of the training data directly motivates SevDiff's design. **Figure 1** shows the distribution of minimum TTC across the 3,581 interaction windows extracted from the smoothed dataset. The left panel shows the full distribution, capped at 175 s: it is heavily right-skewed, with a large concentration below 10 s and a long sparse tail extending to very high TTC values, confirming that the overwhelming majority of recorded interactions are routine. The right panel zooms into the high-severity tail at or below 5.0 s, which is the upper bound of SevDiff's conditioning sweep. Three vertical lines mark the key thresholds: 104 windows fall at or below 1.0 s (2.9% of the 3,581 windows with defined min-TTC), 164 at or below 2.0 s (4.6%), and 237 at or below 3.0 s (6.6%). Only 500 windows (14.0%) fall within the 0–5 s conditioning sweep at all. Three consequences follow from this structure: first, any model trained on this distribution without severity conditioning will almost never generate TTC ≤ 1.0 s interactions from the natural prior, since those 104 windows represent fewer than 3% of the data; second, the 86.0% of windows outside the 0–5 s range are excluded from conditioning training entirely; third, the 4:1 imbalance between the smallest bin (TTC_1.0s: 19 windows) and the largest (TTC_4.0s: 76 windows) requires explicit rebalancing via inverse-frequency weighted sampling to prevent the more abundant mild-severity bins from dominating gradient updates.

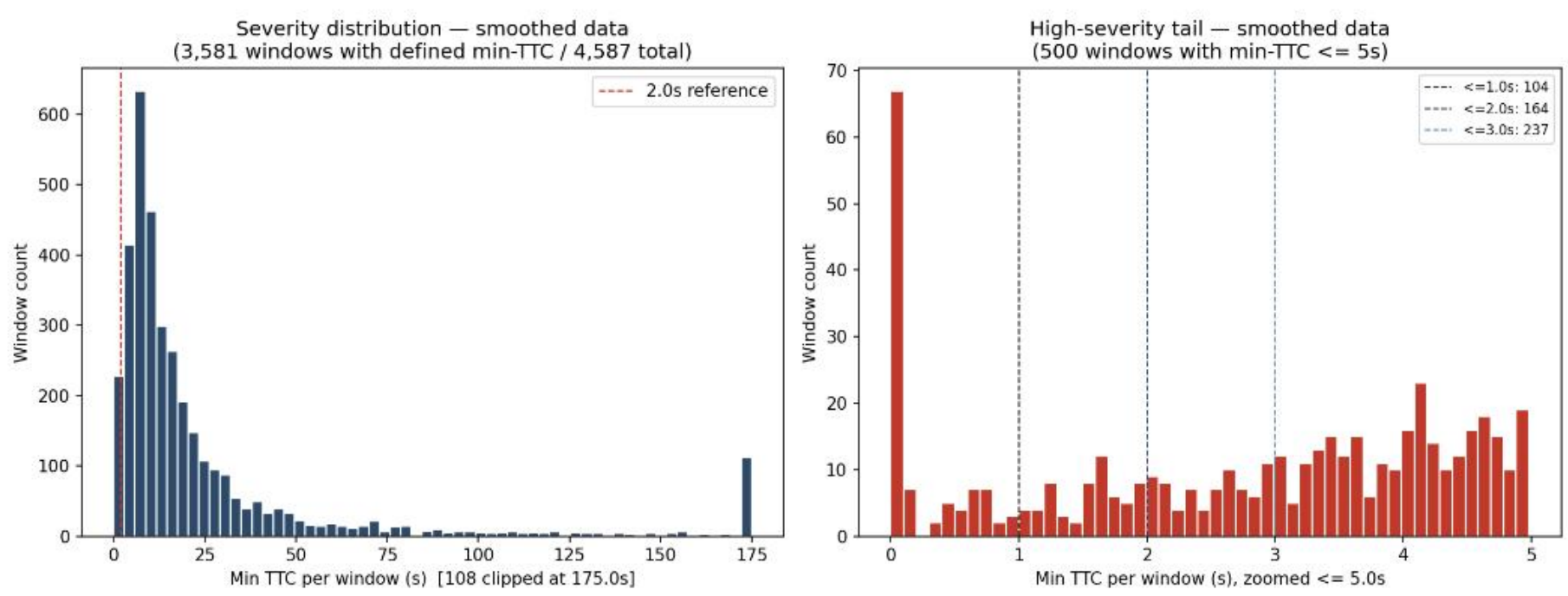

Figure 1: Severity distribution histogram — left panel full distribution, right panel zoomed to ≤5.0 s

### 5.2. Training Convergence

**Figure 2** shows the SevDiff training curves for both Variant A (summary-statistics DDPM, top row) and Variant B (1D convolutional encoder DDPM, bottom row). Each row shows the full 500-epoch curve on the left and the final 150 epochs on the right. For Variant A, the loss drops sharply in the first 50 epochs from 1.0 toward 0.3–0.4, then continues to decline more gradually throughout training. The final 150 epochs reveal noisy but stable oscillation around a common mean, with training and validation losses tracking each other closely without divergence — confirming that the model is not overfitting on the 398-window training split. The absence of a prolonged plateau in the full-run panel is meaningful: a denoiser that ignores the conditioning signal $c$ would plateau once it learns the unconditional noise distribution; the continued decline confirms that the severity conditioning is actively being exploited throughout training. The best validation loss of 0.181 for Variant A (achieved near epoch 470) is the checkpoint used for all evaluation results in Section 5.3. Variant B achieves a lower best validation loss of 0.103, reflecting the richer sequence-level representation from the conv encoder; however, as the hit-rate evaluation reveals, this does not translate into better severity conditioning at the difficult end of the sweep, where the conv encoder's latent space fails to maintain adequate conditioning signal for routine TTC values. Variant A is therefore adopted as the primary model throughout.

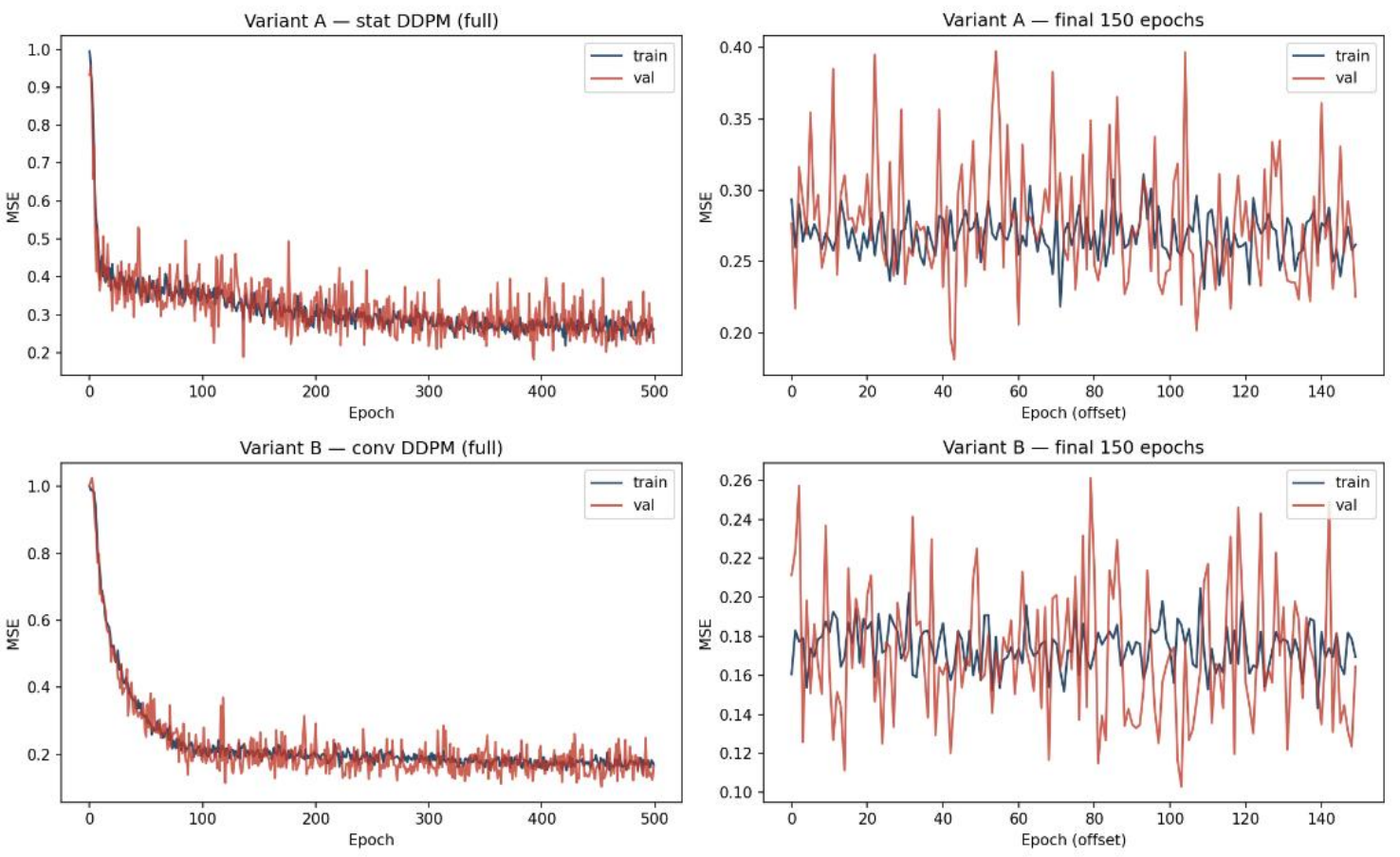

Figure 2: Training curves — Variant A train and validation MSE vs. epoch

### 5.3. Hit-Rate Evaluation

Hit-rate H(δ, r) measures the fraction of N = 300 generated samples whose realized TTC falls within ±δ s of the requested target r. **Table 1 and Figure 3** report results for Variant A (summary-statistics DDPM) across all 10 TTC targets and three error bands.

Specifically, SevDiff achieves 100% hit-rate within ±0.25 s for TTC = 0.5–1.5 s: every generated sample across 900 total generations (300 per target × 3 targets) fell within a quarter-second of the requested severity. For TTC = 2.0–2.5 s, the primary (±0.5 s) hit-rate remains at 97–99%. Degradation becomes pronounced beyond TTC = 3.0 s: 85% at 3.0 s, 53% at 4.0 s, 39% at 5.0 s.

This degradation pattern is interpretable through the conditioning mechanism described in Section 4.3.4. At TTC = 0.5 s, the conditioning ITTC is 2.0 — far from the training prior's center of mass (most interactions have ITTC ≈ 0.2–0.4) — so the term $W_c\varphi(c)$ in equation 5 dominates. At TTC = 5.0 s, the conditioning ITTC of 0.2 is close to the prior's bulk, and the generated distribution is pulled between the conditioning target and the prior. Crucially, the median generated TTC at the 5.0 s target is 4.5 s rather than a random value, confirming that conditioning continues to exert directional influence even where hit-rate is low.

**Table 1.** SevDiff hit-rate results (N = 300 samples per target). Primary metric: ±0.5 s (bold).

| TTC target | Gen. median (s) | ±0.25 s | ±0.5 s (primary) | ±1.0 s |
|---|---|---|---|---|
| 0.5s | 0.4925s | 100.0% | **100.0%** | 100.0% |
| 1.0s | 0.9773s | 100.0% | **100.0%** | 100.0% |
| 1.5s | 1.4631s | 100.0% | **100.0%** | 100.0% |
| 2.0s | 1.9776s | 90.3% | **99.3%** | 100.0% |
| 2.5s | 2.4636s | 74.0% | **97.3%** | 100.0% |
| 3.0s | 2.8974s | 51.3% | **85.0%** | 98.7% |
| 3.5s | 3.3039s | 40.7% | **69.3%** | 95.7% |
| 4.0s | 3.7243s | 27.0% | **53.3%** | 89.0% |
| 4.5s | 4.2457s | 25.7% | **45.7%** | 81.3% |
| 5.0s | 4.546s | 18.3% | **38.7%** | 68.3% |

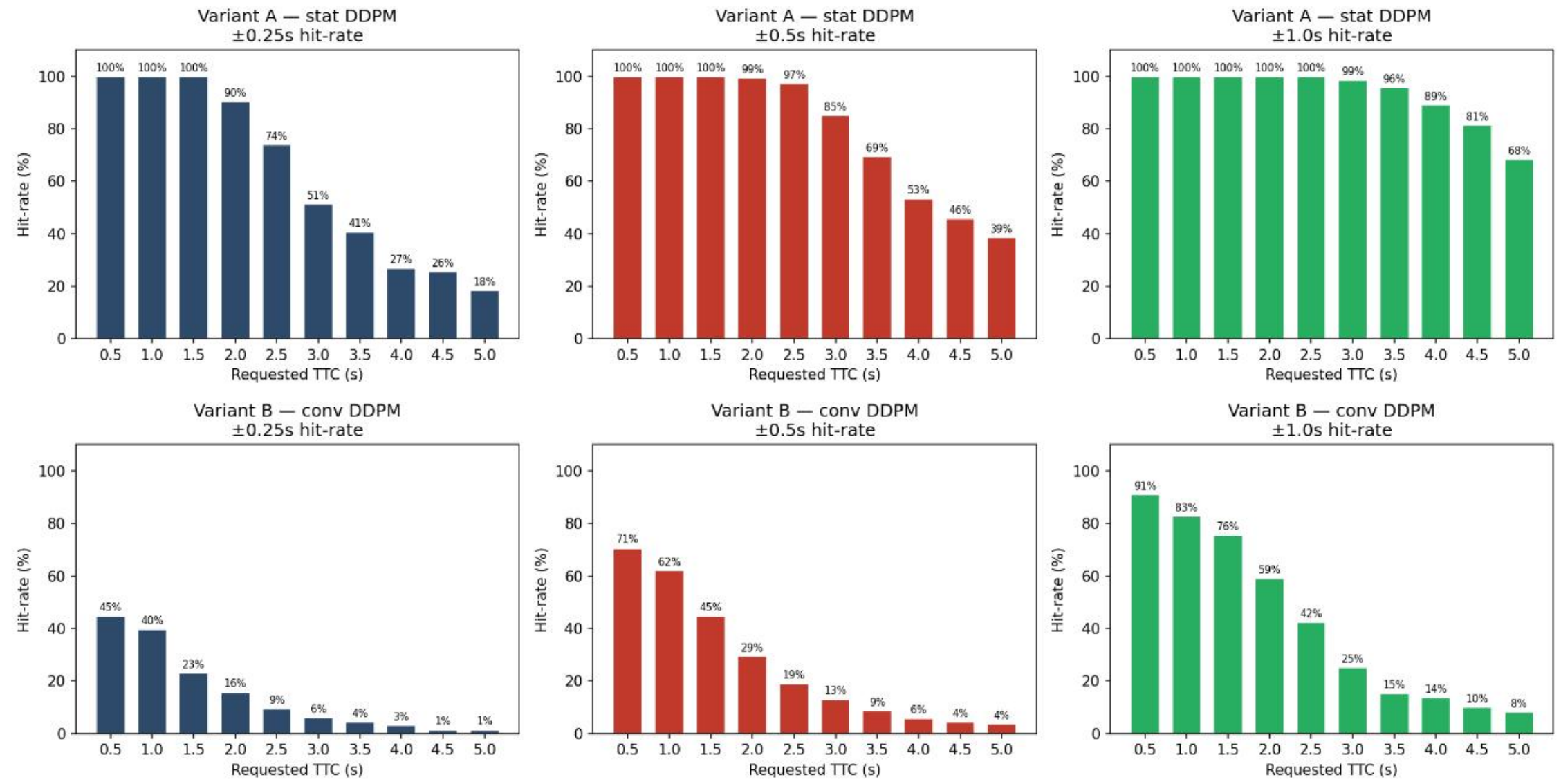

Figure 3: Hit-rate curves — Variant A, three error-band panels

### 5.4. Feature Realism

A distributional realism assessment evaluated whether the 12 generated kinematic features are physically plausible and statistically consistent with the real training distribution. **Figure 4** shows the real versus generated distributions across all 2 features for 1,000 generated samples (100 per target). (See Appendix). **Table 2** summarises per-feature out-of-range (OOR) rates and Kolmogorov-Smirnov statistics.

Mean follower and leader speeds had OOR rates of 0.4% and 0.7% respectively. Mean relative speed and max ITTC had 0.0% OOR. Speed variability features showed rates of 4.7% and 4.2%, attributable to the DDPM occasionally generating slightly negative variance proxy values for brief interactions. KS statistics were below 0.15 for eight of twelve features. The two highest KS statistics were n_frames_norm (KS = 0.311) and max_ittc (KS = 0.255), but neither reflects a generation defect: n_frames_norm under-represents the longest windows because the DDPM generates durations slightly shorter than the training mean, and the broader max_ittc distribution is a direct consequence of sampling across all 10 conditioning targets simultaneously — a generated pool spanning TTC = 0.5–5.0 s will by construction cover more ITTC range than the routine-dominated real distribution.

Note that 54 samples (5.4%) had mean ITTC slightly exceeding max ITTC (physically impossible) and 35 samples (3.5%) had negative minimum headway, both arising from unconstrained generation in the continuous feature space. These violations affect a small fraction of samples and do not alter the hit-rate result; they are addressed as a future direction in Section 6.1.

**Table 2.** Feature realism summary for 1,000 generated samples. OOR = out-of-range. KS = Kolmogorov-Smirnov statistic vs. real training distribution.

| ID | Feature | Real mean | Gen. mean | OOR % | KS | Note |
|---|---|---|---|---|---|---|
| 0 | mean_spd_f (m/s) | 11.511 | 11.130 | 0.4% | 0.120 | |
| 1 | std_spd_f (m/s) | 0.998 | 0.823 | 4.7% | 0.103 | |
| 2 | mean_spd_l (m/s) | 9.703 | 9.271 | 0.7% | 0.121 | |
| 3 | std_spd_l (m/s) | 0.919 | 0.788 | 4.2% | 0.086 | |
| 4 | mean_gap (m) | 14.179 | 11.681 | 1.4% | 0.129 | |
| 5 | min_gap (m) | 9.602 | 7.654 | 3.5% | 0.157 | * |
| 6 | std_gap (m) | 2.761 | 2.381 | 3.1% | 0.072 | |
| 7 | max_ittc | 0.414 | 0.600 | 0.0% | 0.255 | † |
| 8 | mean_ittc | 0.175 | 0.232 | 1.3% | 0.159 | |
| 9 | mean_rel_spd (m/s) | 1.808 | 1.837 | 0.0% | 0.096 | |
| 10 | duration_s | 8.486 | 6.590 | 4.4% | 0.117 | |
| 11 | n_frames_norm | 0.696 | 0.629 | 2.0% | 0.311 | † |

* 3.5% of samples have negative minimum headway (physically impossible; arises from unconstrained generation). † Expected deviation: max_ittc is intentionally diverse across all 10 conditioning targets; n_frames_norm systematically under-represents the longest windows.

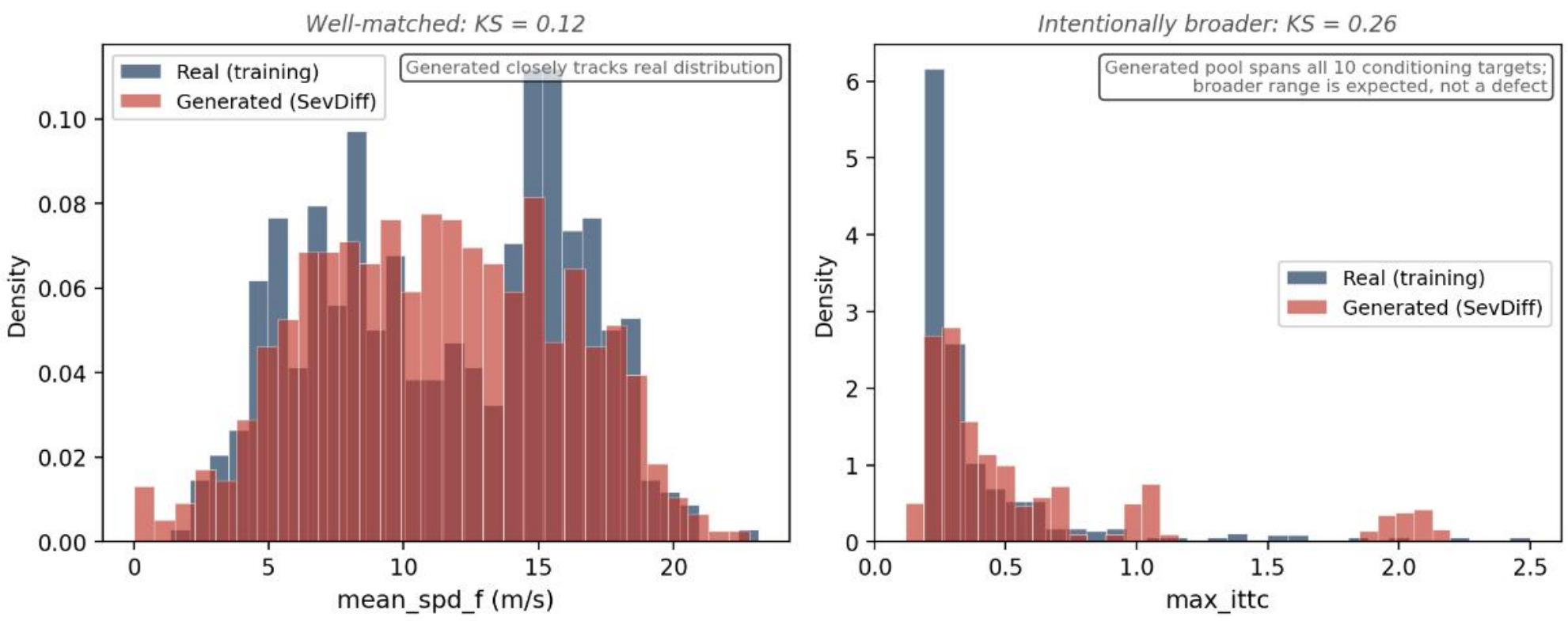

Figure 4: Real vs. generated feature distributions across all 12 features

### 5.5. Trajectory Reconstruction and Visualization

SevDiff generates 12 summary statistics, not a per-frame trajectory. To visualise what a generated scenario looks like in kinematic terms, a deterministic reconstruction step is applied: given the generated summary statistics, the simplest smooth trajectory consistent with them is constructed under five documented assumptions (A1–A5), where A4 is the only hard physical constraint — the gap at the conflict instant is set so that the ITTC formula exactly reproduces the generated ITTC_max target. This reconstruction is not a learned SevDiff capability and must not be interpreted as such; it serves visualisation purposes only.

**Figure 5** shows a side-by-side comparison of one real interaction window (left) and one reconstructed generated scenario (right). The real example is window 103_92_3 from the TTC_0.5s severity bin: the follower (solid) and leader (dashed) position lines converge throughout the window, closing from an initial gap of approximately 9 m to a minimum gap of 4.9 m at the final observed instant, where the annotated TTC is 0.70 s. The ITTC strip below rises steadily from near zero to 1.4 $s^{-1}$ at the conflict instant, colour-shifting from green through yellow to red as the danger increases. This is a genuine, recorded near-conflict event from the naturalistic weaving-section data. The right panel shows the deterministic reconstruction for a TTC = 1.0 s generation request. The follower line closes on the leader progressively, reaching a minimum gap of 4.1 m at the conflict instant (marked by the dotted vertical line), with a realised TTC of 1.04 s — within 0.04 s of the request. The ITTC strip below rises to a peak of approximately 0.98 $s^{-1}$, consistent with the generated ITTC_max of 0.965. The reconstruction note (lower right of the right panel) explicitly states this is a deterministic visualisation under assumptions A1–A5 (Section 5.5), not a learned SevDiff trajectory output. Together, the two panels illustrate that the generated summary statistics produce kinematic configurations structurally similar to real dangerous interactions: closing follower, narrow gap at the conflict instant, and a severity profile matching the requested TTC.

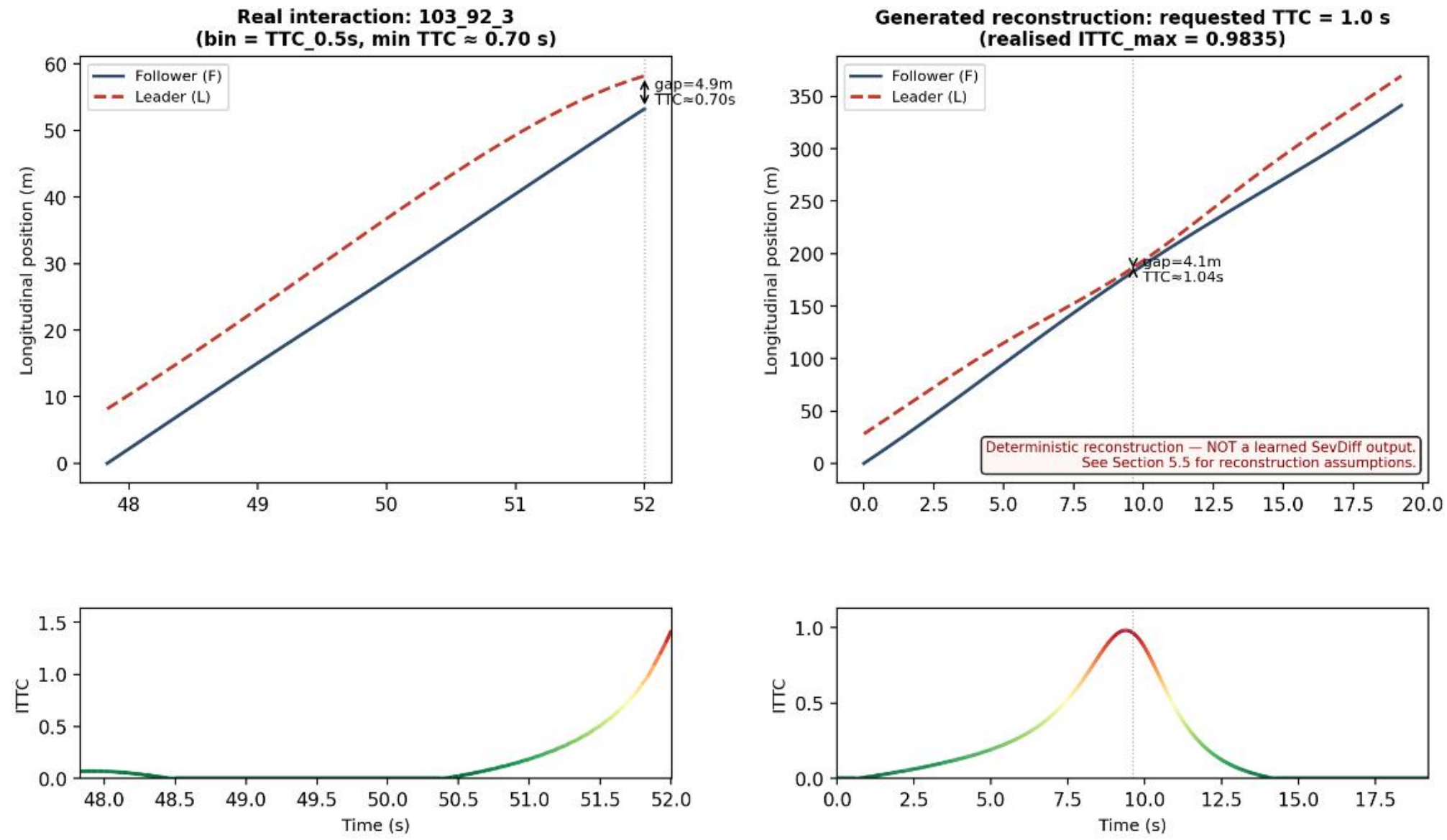

**Figure 5: Position-vs-time diagrams each with ITTC-vs-time strip below**

### 5.6. Nearest-Neighbor Plausibility Yardstick

Feature-space plausibility was assessed by comparing each generated scenario against the real training distribution using a 109,278-pair real-to-real distance yardstick. In the normalized 12-dimensional feature space, the real-to-real distance distribution has a median of 4.153 and interquartile range [3.089, 5.446]. The TTC = 3.0 s generated examples nearest real match (normalized distance 1.294) falls below the 5th percentile of this distribution, placing it closer to a real same-severity scenario than 98.5% of all real-to-real comparisons. The TTC = 1.0 s generated example (distance 3.258) sits at the 71st percentile of all real-to-real pairs — solidly within the normal range.

Within-bin comparisons provide the apple-to-apple check. The TTC = 3.0 s generated example is closer to a real TTC ≈ 3.0 s interaction than 94.6% of same-severity real pairs (within-bin median 3.425, n = 43 windows, 903 pairs). The TTC = 1.0 s generated example sits at the 70th percentile of within-bin distances (within-bin median 4.338, n = 19 windows, 171 pairs). Note that the TTC = 1.0 s same-bin comparison is less statistically stable due to the smaller bin size — a direct consequence of the natural sparsity of genuine near-collision events in the training data that SevDiff was designed to address.

## 6. Discussion

The central result — 100% hit-rate at TTC ≤ 1.5 s, degrading to 39% at TTC = 5.0 s — is precisely what the conditioning mechanism in equation 5 predicts. The conditioning signal ITTC_target is proportional to 1/TTC_target, so dangerous targets map to large ITTC values that sit far from the training prior's center of mass, making them easy to locate. Routine targets map to small ITTC values close to the prior's bulk, where the prior and the conditioning signal compete. The hit-rate curve is therefore not a binary report of success or failure but a precision characterization of the generator across the severity spectrum — a new form of result for the trajectory generation literature.

Relative to prior work, the distinction is both architectural and evaluative. AdvDiffuser [11] achieves dangerous generation by maximizing a collision-probability objective; it answers the question "can I make this dangerous?" SevDiff answers the different question: "can I make this precisely as dangerous as requested, and by how much does precision degrade as the request moves toward the dangerous tail?" The hit-rate metric is the instrument for the second question. The nearest-neigbour yardstick in Section 5.6 further shows that SevDiff's generated feature vectors are not statistical outliers relative to the real training distribution — the TTC = 3.0 s generated example is closer to a real same-severity interaction than 94.6% of same-severity real pairs, grounding the conditioning claim in demonstrated proximity to real data.

### 6.1. Limitations

Three limitations should be stated as mechanistic observations. First, note that SevDiff is trained on 468 windows from a single weaving-section geometry (SQM-W-1); generalization to other road types, sensor configurations, and cultural driving patterns is not tested here, and results should not be assumed to transfer without retraining on geometry-matched data. Second, note that the model operates in an unconstrained continuous feature space: 3.5% of generated samples produce negative minimum headway and 5.4% produce mean ITTC exceeding max ITTC, both physically impossible. These violations are not resolved by the current architecture — they require either post-generation clipping or an inequality-constrained generation objective. Third, note that the 12-feature summary representation enables strong conditioning with small data but discards temporal structure; the reconstruction in Section 5.5 is deterministic and visualization-only, not a learned trajectory synthesis capability.

### 6.2. Future Directions

Three future directions follow, ordered by proximity to the current work. (i) *Constrained generation*: the 3.5% negative $min_{gap}$ and 5.4% mean-ITTC-exceeds-max-ITTC violation rates are known defects in the unconstrained feature space; adding physical inequality constraints to the generation objective would close these gaps without requiring architectural change. (ii) *Multi-dataset training*: SevDiff is currently limited to one weaving-section geometry; training on multiple datasets with varied road types, density conditions, and geographic driving cultures would test whether the hit-rate curve generalizes beyond SQM-W-1. Future work should also explore applying SevDiff to closed-loop trajectory datasets such as nuPlan [40,41] or Waymo Open Dataset [42], both of which provide richer multi-agent interaction context and HD map geometry, enabling scene-conditioned severity generation where spatial road structure is available alongside the kinematic features. (iii) *VectorNet-style scene encoding*: replacing the 12-feature summary representation with a polyline-based spatial graph would enable full per-frame trajectory generation, making the deterministic reconstruction step unnecessary and the visualization grounded in a learned output.

## 7. Conclusions

SevDiff achieves 100% hit-rate within ±0.5 s for TTC targets of 0.5–1.5 s and 97–99% at 2.0–2.5 s, demonstrating that conflict-severity conditioning in a denoising diffusion model can be made precise and measurable. The model is trained on 468 naturalistic interaction windows from the UTE SQM-W-1 expressway dataset and evaluated through a hit-rate metric that directly quantifies whether the requested severity was achieved — a metric that does not appear in the existing trajectory generation literature. The hit-rate degradation from 100% at TTC = 1.5 s to 39% at TTC = 5.0 s is physically interpretable as the strength of the conditioning signal relative to the training prior, making it a diagnostic of the generator rather than a simple performance claim. Generated feature vectors are physically plausible (maximum OOR 4.7% across 12 features) and statistically close to real data (TTC = 3.0 s generated example closer to a real same-severity scenario than 94.6% of same-severity real pairs). Future work will extend SevDiff through constrained generation, multi-dataset training including closed-loop datasets with HD map support, and VectorNet-style scene encoding.

**Author Contributions:** Conceptualization, methodology, software, validation, formal analysis, investigation, data curation, and writing (original draft, review and editing), E.S.L. The author has read and agreed to the published version of the manuscript.

**Funding:** This research received no external funding.

**Institutional Review Board Statement:** Not applicable.

**Informed Consent Statement:** Not applicable.

**Data Availability Statement:** The UTE dataset is available from the Ubiquitous Traffic Eye project. Code will be made available upon request.

**Conflicts of Interest:** The authors declare no conflicts of interest.

**Abbreviations**

The following abbreviations are used in this manuscript:

| | |
|---|---|
| ADAS | Advanced Driver Assistance System |
| DDPM | Denoising Diffusion Probabilistic Model |
| FCW | Forward Collision Warning |
| ITTC | Inverse Time-to-Collision |
| KS | Kolmogorov-Smirnov |
| MLP | Multi-Layer Perceptron |
| MSE | Mean Squared Error |
| ODD | Operational Design Domain |
| OOR | Out-of-Range |
| POT | Peak-Over-Threshold |
| SevDiff | Severity-Conditioned Diffusion |
| SOTIF | Safety of the Intended Functionality |
| SSM | Surrogate Safety Measure |
| TTC | Time-to-Collision |
| UAV | Unmanned Aerial Vehicle |

UTE Ubiquitous Traffic Eye

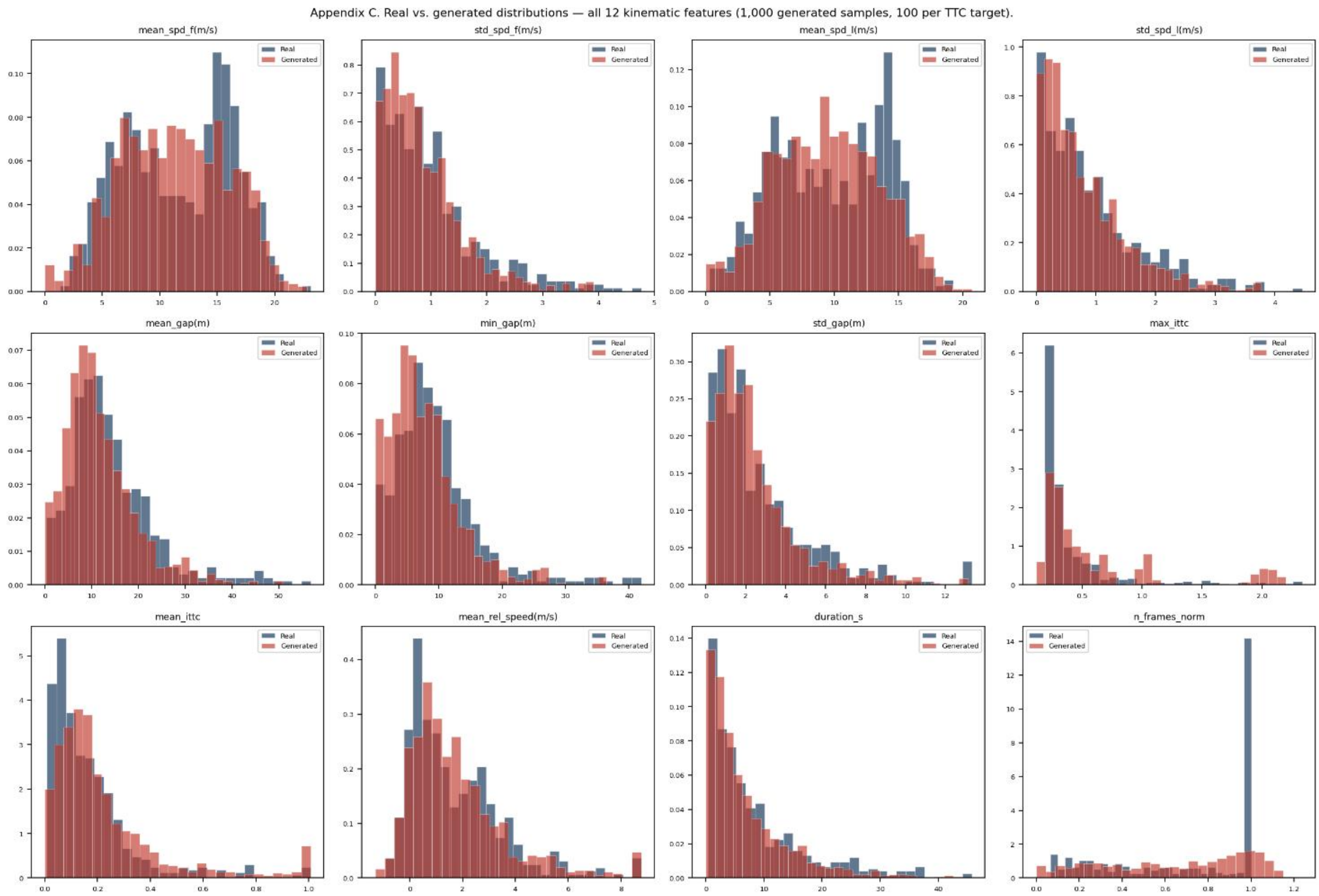

Figure 4: Real vs. generated feature distributions across all 12 features

## Appendix A: SevDiff Training Algorithm

The following pseudocode summarises the SevDiff training procedure. Section references indicate where each step is described in the main text.

**Algorithm A1: SevDiff — Training Pipeline**

INPUT (Section 4.3.1)
    Raw trajectory data: frenet.csv — 1,041 vehicles, 822,712 rows at 24 Hz
STEP 1 Data diagnostics and smoothing (Section 4.3.1)
    Run internal consistency checks D1–D7: confirm position-speed, leader-distance,
     leader-follower reciprocity, speed non-negativity, and frame continuity
    Confirm D2 acceleration scheme: centered difference, RMSE = 0.134 m/s²
    Savitzky-Golay smooth speed (window=7, poly=3) and lateral distance (window=11, poly=3)
    Recompute acceleration from smoothed speed using centered differences
    Apply D5 filter: remove 21 rows with non-positive leader headway
STEP 2 ITTC computation and pair extraction (Section 4.3.2)
    FOR each row with a valid leader link:
        Compute ITTC(t) = max(v_f − v_l, 0) / d
    Segment (follower, leader) pairs into continuous interaction windows
    Compute ITTC_max = max over window W of ITTC(t)
    Extract 12 summary features per window
    Retain windows with ITTC_max > 0 and ≥ 10 frames (N = 3,581 windows)
STEP 3 Severity binning (Section 4.3.3)
    Apply Peak-Over-Threshold analysis: upper bound = TTC 5.0 s
    Bin into 10 strata: TTC 0.5 s–5.0 s (±0.25 s half-width)
    Training set: 468 windows across 10 bins (range: 19–76 per bin)
    Apply inverse-frequency weighted random sampling across bins
STEP 4 Conditional DDPM training (Section 4.3.4)
    Compute feature mean/std on training split; normalize $x_0$
    Initialise $\varepsilon_\theta$: 4-layer MLP, 256 hidden units, SiLU activations
    Set β schedule: linear from $10^{-4}$ to 0.02 over T = 300 steps
    FOR epoch = 1 to 500:
        Sample ($x_0$, c = ITTC_max) from weighted training set
        Sample t ~ Uniform{1, …, 300}
        Compute $x_t$ = sqrt($\bar{\alpha}$_t) · $x_0$+ sqrt(1 − $\bar{\alpha}$_t) · ε
        Compute h = W_x · x_t + W_t · φ(t) + W_c · φ(c)
        Compute loss L = || ε − MLP(h) ||^2
        Update θ via Adam, lr = $3\times10^{-4}$, cosine annealing to $10^{-6}$
    Save best checkpoint on held-out 15% validation split
OUTPUT Trained denoising network $\varepsilon_\theta$ with feature normalisation statistics

## Appendix B: SevDiff Inference Algorithm

The following pseudocode summarises hit-rate evaluation and optional trajectory reconstruction. Section references indicate where each step is described in the main text.

**Algorithm B1: SevDiff — Inference and Evaluation**

```
INPUT  (Section 4.3.5)
    Requested TTC target: r ∈ R_sweep = {0.5, 1.0, …, 5.0} s
    Trained ε_θ and feature normalisation statistics from Appendix A
STEP 1  Severity conditioning setup  (Section 4.3.5)
    Convert TTC target to ITTC conditioning value: c = 1 / r
STEP 2  Reverse diffusion  (Section 4.3.4, equation 4)
    FOR sample i = 1 to N = 300:
        Initialize: x_T ~ N(0, I)
        FOR t = T down to 1:
            Compute h = W_x · x_t + W_t · φ(t) + W_c · φ(c)
            Predict noise: ε_pred= MLP(h)
            Compute posterior mean μ(x_t, ε_pred)
            x_{t-1} = μ + σ · noise  (if t > 1),  else x_0 = μ
        Denormalize x_0 to recover raw 12-feature vector
        Read ITTC_gen = feature[7] (ITTC_max)
        Convert: TTC_gen,i = 1 / ITTC_gen
STEP 3  Hit-rate evaluation  (Section 4.3.5, equation 7)
    FOR each error band δ ∈ {0.25, 0.5, 1.0} s:
        H(δ, r) = (1/N) · Σ 1[ |TTC_gen,i − r| ≤ δ ]
    Report H(δ, r) across all r ∈ R_sweep  → hit-rate curve
STEP 4  Nearest-neigbour plausibility check  (Section 5.6)
    Normalize generated features using training mean/std
    Compute L2 distance to all real training windows
    Report percentile rank vs. real-to-real distance distribution
STEP 5  Trajectory reconstruction for visualization  (Section 5.5, OPTIONAL)
    Apply deterministic reconstruction (assumptions A1–A5):
        A1: frame rate = 0.0417 s, n_frames from generated duration_s
        A2: speed profiles = smooth sinusoids at generated mean ± std
        A3: conflict instant placed at window midpoint
        A4: gap at conflict instant set to enforce generated ITTC_max exactly
        A5: lateral position held constant (no lane change depicted)
    Output: position-vs-time diagram with ITTC-vs-time strip
    Note: this step is NOT a learned SevDiff output; visualization only
OUTPUT  Hit-rate table H(δ, r), plausibility yardstick, optional trajectory plots
```